\documentclass[10pt,twocolumn,letterpaper]{article}

\usepackage{iccv}
\usepackage{times}
\usepackage{epsfig}
\usepackage{graphicx}
\usepackage{amsmath}
\usepackage{amssymb}
\usepackage{subfigure}
\usepackage{booktabs}
\usepackage{multicol}
\usepackage{multirow}

\usepackage[breaklinks=true,bookmarks=false]{hyperref}

\iccvfinalcopy 

\def\iccvPaperID{****} 

\ificcvfinal\fi

\begin{document}

\title{HRegNet: A Hierarchical Network for\\Large-scale Outdoor LiDAR Point Cloud Registration}
\author{Fan Lu\textsuperscript{1}, Guang Chen\textsuperscript{1,}\thanks{Corresponding author: guangchen@tongji.edu.cn}, Yinlong Liu\textsuperscript{2}, Lijun Zhang\textsuperscript{1}, Sanqing Qu\textsuperscript{1}, Shu Liu\textsuperscript{3}, Rongqi Gu\textsuperscript{4}\\
\textsuperscript{1}Tongji University, \textsuperscript{2}Technische Universität München, \textsuperscript{3}ETH Zurich, \textsuperscript{4}Westwell lab\\
{\tt\small \{lufan,guangchen,tjedu\_zhanglijun,2011444\}@tongji.edu.cn}\\
{\tt\small Yinlong.Liu@tum.de, liush@ethz.ch, rongqi.gu@westwell-lab.com}}


\maketitle
\ificcvfinal\thispagestyle{empty}\fi

\begin{abstract}
   Point cloud registration is a fundamental problem in 3D computer vision. Outdoor LiDAR point clouds are typically large-scale and complexly distributed, which makes the registration challenging. In this paper, we propose an efficient hierarchical network named HRegNet for large-scale outdoor LiDAR point cloud registration. Instead of using all points in the point clouds, HRegNet performs registration on hierarchically extracted keypoints and descriptors. The overall framework combines the reliable features in deeper layer and the precise position information in shallower layers to achieve robust and precise registration. We present a correspondence network to generate correct and accurate keypoints correspondences. Moreover, bilateral consensus and neighborhood consensus are introduced for keypoints matching and novel similarity features are designed to incorporate them into the correspondence network, which significantly improves the registration performance. Besides, the whole network is also highly efficient since only a small number of keypoints are used for registration. Extensive experiments are conducted on two large-scale outdoor LiDAR point cloud datasets to demonstrate the high accuracy and efficiency of the proposed HRegNet. The project website is \url{https://ispc-group.github.io/hregnet}.

\end{abstract}

\section{Introduction}
Point cloud registration aims to estimate the optimal rigid transformation between two point clouds, which is a fundamental problem in 3D computer vision and plays an important role in many applications such as robotics~\cite{pomerleau2015review} and autonomous driving~\cite{nagy2018real}.

Iterative Closest Point (ICP) \cite{icp} is the best-known method to solve point cloud registration problem. However, ICP highly relies on the initial guesses of the transformation for iteration and can easily get stuck into local minimum due to the non-convexity of the problem. Several variants of ICP \cite{rosen2019se,maron2016point,yang2015go} have been proposed to achieve global optimal estimation, however, are typically time-consuming for large-scale point clouds.

\begin{figure}
	\centering
	\includegraphics[width=0.44\textwidth]{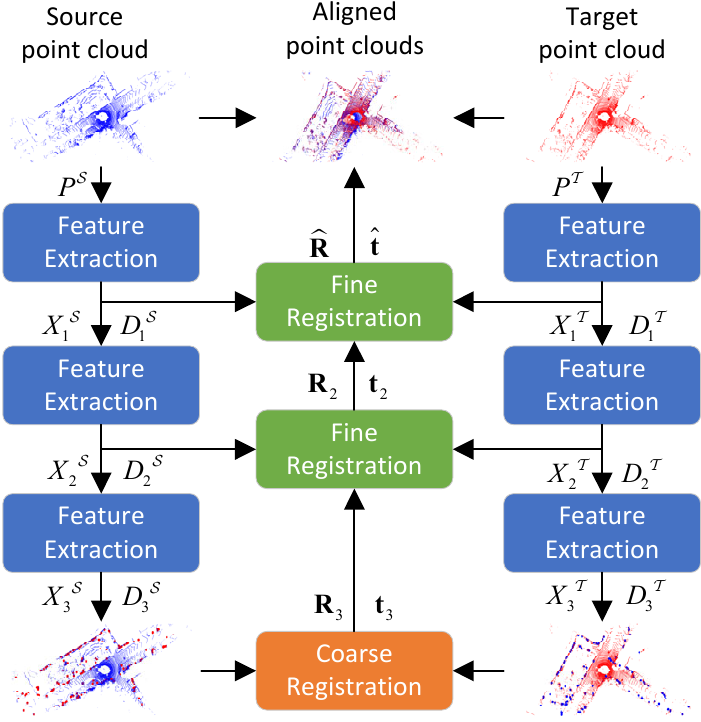}
	\caption{Network architecture of the proposed HRegNet. The point clouds $P$ are hierarchically downsampled to small sets of keypoints $X$ and descriptors $D$. We perform coarse registration in the bottom layer to leverage the reliable features for keypoints matching and fine registration is followed to refine the transformation by exploiting the precise position information in upper layers.}

	\label{fig:overall}
	\vspace{-5mm}
\end{figure}

Recently, deep learning has achieved great success in numerous 3D computer vision tasks such as 3D object detection and semantic segmentation \cite{guo2020deep}. There also emerge a number of deep learning-based methods for point cloud registration. However, existing methods are mostly designed for object-level point clouds \cite{wang2019deep,aoki2019pointnetlk,wang2019prnet,yuan2020deepgmr,idam} or indoor point clouds \cite{choy2020deep, huang2020feature, pais20203dregnet,gojcic2020learning}. Compared to object-level or indoor point clouds, outdoor LiDAR point clouds typically have higher sparsity, larger spatial range and a more complex and variable distribution, which makes the registration intractable. Consequently, existing learning-based methods are either unreliable or time-consuming to be applied to outdoor LiDAR point cloud registration.

In this paper, we aim to provide an accurate, reliable and efficient network for large-scale outdoor LiDAR point cloud registration. Inspired by the success of learning-based 3D features on LiDAR point cloud registration \cite{li2019usip,bai2020d3feat,lu2020rskdd,yew20183dfeat,choy2019fully}, we propose a hierarchical keypoint-based point cloud registration network named HRegNet. The overall structure is displayed in Fig.~\ref{fig:overall}. We hierarchically downsample the point clouds to multiple small sets of keypoints and descriptors for registration. Intuitively, as the layer goes deeper, the information of a single keypoint increases, which makes the descriptors more reliable for keypoints matching, however, the increasing sparsity of keypoints may also cause larger position error of corresponding keypoints. Based on the above consideration, the network starts with coarse registration in the bottom layer by globally matching keypoints in descriptor space to leverage the reliable features. Then the coarse transformation is refined by fine registration in upper layers based on local matching in spatial neighborhoods, which exploits the precise position information in shallower layers. Besides, since only a small number of keypoints are used for registration, the network has high efficiency and can be applied in applications requiring real-time performance, such as autonomous driving.

\begin{figure}

    \centering
	\subfigure[Bilateral consensus.]{
	\includegraphics[height=0.12\textwidth]{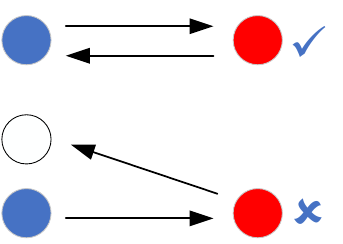}
	\label{fig:bilateral}
	}
	\subfigure[Neighborhood consensus.]{
	\includegraphics[height=0.12\textwidth]{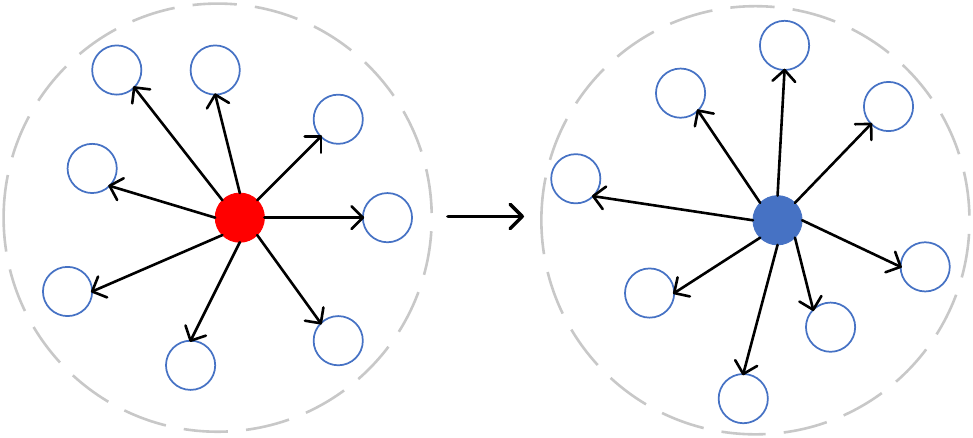}
	\label{fig:neighbor}
	}
	\caption{(a) Bilateral consensus: Two corresponding keypoints should be the nearest neighbor in descriptor space of each other. (b) Neighborhood consensus: Spatial neighborhoods of two corresponding keypoints should also be similar.}
	\label{fig:consensus}
	\vspace{-4mm}
\end{figure}

Although the keypoints in the bottom layer have reliable features, possible error of descriptors may lead to a considerable number of mismatches. To improve the robustness and accuracy of registration, we present a learning-based correspondence network to generate corresponding keypoints and reject unreliable correspondences. Here we introduce two important concepts for keypoints matching, namely bilateral consensus and neighborhood consensus. Bilateral consensus, as illustrated in Fig.~\ref{fig:bilateral}, means that a pair of corresponding keypoints should be the nearest neighbor in descriptor space of each other from both sides. As shown in Fig.~\ref{fig:neighbor}, neighborhood consensus indicates that the neighboring keypoints of two corresponding keypoints should also have high similarity. Notably, bilateral consensus and neighborhood consensus have been successfully applied in many cases (\emph{e.g.}, estimate image dense correspondences~\cite{rocco2018}). To effectively incorporate them into the learning-based registration pipeline, we design novel similarity features based on bi-directional similarity of descriptors and an attention-based neighbor encoding module, which significantly improves the registration performance.

To evaluate the proposed HRegNet, extensive experiments are performed on two large-scale outdoor LiDAR point cloud datasets, namely KITTI odometry dataset \cite{geiger2012we} and NuScenes dataset \cite{caesar2020nuscenes}. The results demonstrate that the proposed method significantly outperforms existing methods in terms of both accuracy and efficiency.

In summarize, our main contributions are as follows:
\begin{itemize}
\setlength{\itemsep}{0pt}
\setlength{\parsep}{0pt}
\setlength{\parskip}{0pt}
    \item We propose a novel point cloud registration network named HRegNet, which achieves state-of-the-art performance with high computational efficiency.
    \item The hierarchical paradigm well combines the strengths of keypoints and descriptors in shallower and deeper layers to achieve precise and robust registration.
    \item We design novel similarity features, which effectively incorporate bilateral consensus and neighborhood consensus into the registration pipeline and significantly improve the registration performance.
\end{itemize}

\section{Related works}

We briefly review the related works in two aspects: classical and learning-based point cloud registration methods.

\paragraph{Classical point cloud registration:} Iterative closest point (ICP) \cite{icp} is the best-known algorithm for point cloud registration, which iteratively finds the closest point and updates the transformation by solving a least square problem. However, ICP is a local registration algorithm and can easily get stuck into local minimum. Several variants \cite{rosen2019se,maron2016point,yang2015go} aim to break the limitation of ICP. Go-ICP \cite{yang2015go} uses a Branch-and-Bound (BnB) algorithm to search a global optimal solution. Several methods also try to extract features from point clouds for registration \cite{flint2007thrift,rusu2009fast,sipiran2011harris,tombari2010unique,johnson1999using}. For example, Fast Point Feature Histogram (FPFH) \cite{rusu2009fast} builds an oriented histogram using pairwise geometric properties. A comprehensive review of handcrafted features in 3D point clouds can be found in \cite{guo2016comprehensive}. After feature extraction, \textit{RANdom SAmple Consensus} (RANSAC) \cite{fischler1981random} is commonly used for robust feature matching by randomly sampling small subsets of correspondences and then finding optimal correspondences for registration.

\paragraph{Learning-based point cloud registration:} PointNetLK is a pioneering work of learning-based point cloud registration \cite{aoki2019pointnetlk}. It performs registration by combines PointNet \cite{qi2017pointnet} and Lucas \& Kanade algorithm \cite{lucas1981iterative} into a single trainable recurrent deep neural network. Deep Closest Point (DCP) \cite{wang2019deep} is a well-known learning-based point cloud registration network. It uses a transformer network to predict soft matching between point clouds and provides a differentiable Singular Value Decomposition (SVD) layer to calculate transformation. IDAM \cite{idam} utilizes an iterative distance-aware similarity matrix convolution module for pairwise points matching. However, the above methods are basically designed for object-level point clouds and not applicable to complex large-scale LiDAR point clouds.

Recently, there emerge several learning-based methods for indoor point cloud registration. 3DRegNet \cite{pais20203dregnet} proposes to use deep network to directly regress the transformation. Feature-metric registration \cite{huang2020feature} aims to solve the registration problem from a different perspective. It performs registration by minimizing a feature-metric projection error without correspondences rather than minimizing commonly used geometric error. Gojcic \etal mainly focus on the registration of multiview 3D point clouds \cite{gojcic2020learning}. Deep Global Registration (DGR) \cite{choy2020deep} proposes to use a learning-based feature named Fully Convolutional Geometric Features (FCGF) \cite{choy2019fully} to perform registration. A 6D convolutional network \cite{choy20194d} is adopted to predict a likelihood for each correspondence. DGR achieves state-of-the-art performance in indoor point cloud registration. DeepVCP \cite{lu2019deepvcp} is a method designed for LiDAR point cloud registration. It proposes to use virtual points to construct correspondences. However, the keypoints matching in DeepVCP is performed only in local 3D coordinate space, which makes the method can be basically applied to local registration problem.

\section{Methodology}

Given source and target point clouds $P^{\mathcal{S}},P^{\mathcal{T}}\in\mathbb{R}^{N\times 3}$, HRegNet aims to predict the optimal rotation matrix $\hat{\mathbf{R}}$ and translation vector $\hat{\mathbf{t}}$ from source to target point clouds in a coarse-to-fine manner. As shown in Fig.~\ref{fig:overall}, here we adopt a 3-layer implementation. Given a point cloud $P$, we utilize 3 cascaded feature extraction modules to hierarchically downsample the point clouds to multiple small sets of keypoints $X_l\in \mathbb{R}^{M_l\times 3}$, descriptors $D_l\in \mathbb{R}^{M_l\times C_l}$ and also saliency uncertainties $\Sigma_l\in \mathbb{R}^{M_l}$, where $l=\{1,2,3\}$ represents the layer number, $M_l$ is the number of keypoints and $C_l$ is the channel of descriptors. To exploit reliable features of keypoints in the bottom layer, coarse registration is performed by globally matching keypoints in descriptor space to estimate a coarse transformation $\mathbf{R}_3,\mathbf{t}_3$, which is further applied to transform the keypoints in upper layer. After that, we adopt fine registration in layer $l=2$ to refine the coarse transformation. We assume that the coarse transformation can basically align the point clouds, thus, keypoints matching in fine registration is performed locally in spatial neighborhoods. Finally, another fine registration is applied in the top layer to obtain the final estimation $\hat{\mathbf{R}},\hat{\mathbf{t}}$.

\subsection{Feature extraction}

The input of each feature extraction module is the keypoints (or the original point cloud), saliency uncertainties, descriptors and also features of keypoints in previous layer. We firstly adopt Weighted Farthest Point Sampling (WFPS) \cite{zhou2020da4ad,qi2017pointnetplus} to select a set of candidate keypoints. After that, $k$-nearest-neighbor ($k$NN) search is performed to construct clusters centered on the candidate keypoints and a Shared Multi-layer Perceptron (Shared-MLP) \cite{qi2017pointnet} is followed to refine the location of candidate keypoints by predicting attentive weight for each neighboring point in the cluster. Saliency uncertainty is also predicted by applying another Shared-MLP to the cluster. Besides, a descriptor network is designed to extract descriptor from the cluster for each keypoint. Since the feature extraction module is not the main focus of this paper, the detailed network structure is provided in our supplementary material.

\subsection{Coarse registration}
After the keypoints and descriptors are extracted by the feature extraction module, the key problem then is how to find correct correspondences between source and target keypoints. The most commonly used method to match two sets of keypoints is nearest neighbor (NN) search in descriptor space. Although the descriptors in the bottom layer are relatively more reliable, they are not perfect. Thus, the simple NN search-based approach may result in a considerable number of mismatches due to possible errors of descriptors, which will cause a large registration error. To address the above problem, in this paper, we adopt a learning-based correspondence network to match two sets of keypoints in the bottom layer $l=3$ to perform coarse registration.

\subsubsection{Correspondence network}\label{sec:corr}
\begin{figure}
    \centering
    \includegraphics[width=0.45\textwidth]{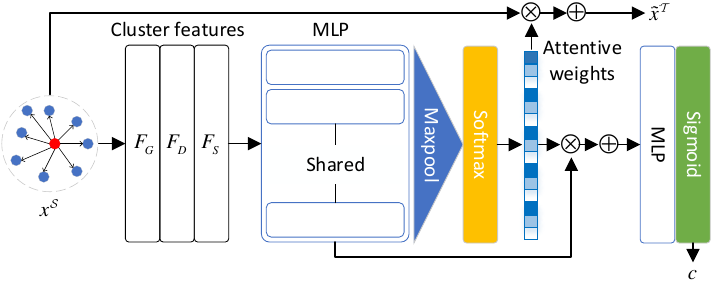}
    \caption{Architecture of correspondence network in coarse registration. The input is the $k$NN cluster of a source keypoint $x^{\mathcal{S}}$ and the features consist of geometric features $F_G$, descriptor features $F_D$ and similarity features $F_S$. The output is the corresponding keypoint $\tilde{x}^\mathcal{T}$ and the confidence score $c$.}
    \label{fig:corres}
    \vspace{-4mm}
\end{figure}

To simplify the formulation, the subscripts $l$ indicating layer number are omitted in this section and we denote the source and target keypoints and descriptors as $X^\mathcal{S},X^\mathcal{T}\in\mathbb{R}^{M\times 3}$ and $D^\mathcal{S},D^\mathcal{T}\in\mathbb{R}^{M\times C}$, respectively. As shown in Fig.~\ref{fig:corres}, for a source keypoint $x^\mathcal{S}$, we firstly perform $k$-nearest-neighbor ($k$NN) search in descriptor space to find $K$ candidate corresponding keypoints in $X^\mathcal{T}$. The $K$ neighboring candidate keypoints $\{x_1^\mathcal{T},\cdots,x_K^\mathcal{T}\}$ and the center keypoint $x^\mathcal{S}$ form a cluster. The features of the cluster consist of three parts: geometric features $F_G$, descriptor features $F_D$ and similarity features $F_S$. $F_G$ is the concatenation of coordinates of the center and neighboring keypoints. Besides, the relative coordinates and distances between neighboring and center keypoints are calculated as additional geometric features. $F_D$ consists of the descriptors of center and neighboring keypoints. In addition, the saliency uncertainties of keypoints are also included in $F_D$. $F_S$ is introduced to incorporate bilateral consensus and neighborhood consensus and will be described in detail in Section~\ref{sec:sim} below. 

The cluster features are firstly passed into a 3-layer Shared-MLP to generate a feature map $\tilde{F}=\{f_1,\cdots,f_K\}$. A max-pool layer and a Softmax function are further applied to predict an attentive weight $w_k^\mathcal{T}$ for each candidate keypoint $x_k^\mathcal{T}$. The estimated corresponding keypoint of $x^\mathcal{S}$ can be represented as the weighted sum of the candidate keypoints. Besides, an attentive feature $\bar{F}$ of the cluster is calculated as the weighted sum of $\tilde{F}$. $\bar{F}$ is further fed into a MLP with a Sigmoid function to predict a confidence score $c$ for this correspondence. Then the confidence score is normalized by $\tilde{c}_i=c_i/\sum_{j=1}^{M}c_j$. As we claimed before, using simple NN search can cause a considerable number of mismatches due to the error of descriptors. Intuitively, the proposed attention-based formulation aims to implicitly assign higher weights to the correct candidate corresponding keypoints. The learning-based paradigm incorporates the geometric features, descriptors and also bilateral consensus and neighborhood consensus to generate accurate correspondences and reject unreliable correspondences using the predicted confidence score $\tilde{c}$. Given the corresponding keypoints and confidence scores, the optimal transformation $\mathbf{R}^{*},\mathbf{t}^{*}$ can be calculated as
\begin{equation}
\label{eq:svd}
    \mathbf{R}^{*},\mathbf{t}^{*}=\mathop{\arg\min}_{\mathbf{R},\mathbf{t}} \sum_{i}^M \tilde{c}_{i}\left\|\mathbf{R}x_i^\mathcal{S}+\mathbf{t}-\tilde{x}_i^\mathcal{T}\right\|_2
\end{equation}
where $x_i^\mathcal{S}$ and $\tilde{x}_i^\mathcal{T}$ are corresponding keypoints, $\tilde{c}_i$ is confidence score and $\left\|\cdot\right\|_2$ denotes $L_2$ norm. Eq.~\ref{eq:svd} can be closed-form solved using weighted Kabsch algorithm \cite{kabsch1976solution}, which has also been derived in detail in \cite{gojcic2020learning}.

\subsubsection{Similarity features}\label{sec:sim}
\textbf{Bilateral consensus:} Based on the $k$NN search, we can only ensure that the searched $K$ candidate keypoints in $X^\mathcal{T}$ are most similar with the keypoint $x^\mathcal{S}$. However, this single directional operation can not guarantee the reverse best similarity of the matching. Intuitively, a correct correspondence should satisfy bilateral consensus, which means that if $x_j^\mathcal{T}$ is the nearest neighbor (in descriptor space) in $X^\mathcal{T}$ of $x_i^\mathcal{S}$, then $x_i^\mathcal{S}$ should also be the nearest neighbor in $X^\mathcal{S}$ of $x_j^\mathcal{T}$.

\begin{figure}
    \centering
    \includegraphics[width=0.45\textwidth]{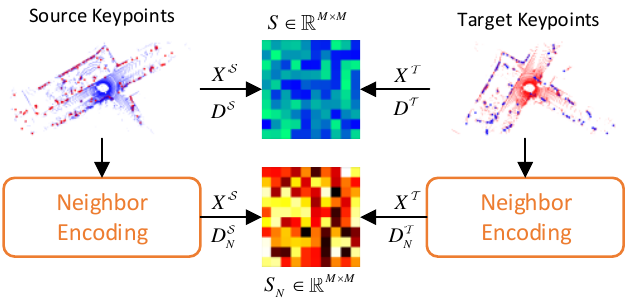}
    \caption{Illustration of similarity matrix. Given source and target keypoints, we calculate the cosine similarity of the descriptors to form $S\in \mathbb{R}^{M\times M}$. The neighbor encoding module is adopted to gather neighborhood information and the similarity matrix $S_N$ is calculated based on the neighbor-aware descriptors $D_N^{\mathcal{S}}$, $D_N^{\mathcal{T}}$.}
    \label{fig:similarity}
    \vspace{-4mm}
\end{figure}

Based on the above consideration, we introduce novel similarity features to take bilateral consensus into consideration. As shown in the top row of Fig.~\ref{fig:similarity}, for each keypoint $x_i^\mathcal{S}$, we calculate the cosine similarity of the descriptor $d_i^\mathcal{S}$ with descriptors of all keypoints in $X^\mathcal{T}$. Consequently, we can obtain a $M\times M$ similarity matrix and an entry $s_{ij}$ of the similarity matrix $S\in \mathbb{R}^{M\times M}$ can be calculated as
\begin{equation}
    s_{ij} = \frac{\langle d_{i}^\mathcal{S},d_{j}^\mathcal{T}\rangle}{\left\|d_{i}^\mathcal{S}\right\|_2\left\|d_{j}^\mathcal{T}\right\|_2}
\end{equation}
where $\langle \cdot,\cdot\rangle$ and $\left\|\cdot\right\|_2$ denote inner product and $L_2$ norm.

After that, we normalize the similarity matrix in two directions to generate two different similarity matrixes $S^f$ (forward matrix) and $S^b$ (backward matrix) as
\begin{equation}
    s_{ij}^f = \frac{s_{ij}}{\max_{m}s_{im}},\quad 
    s_{ij}^b = \frac{s_{ij}}{\max_{m}s_{mj}}
\end{equation}

Then, the similarity features of the cluster are the concatenation of corresponding similarity scores of the candidate keypoints with the center keypoint in forward and backward similarity matrix $S^f$ and $S^b$. Take a pair of candidate corresponding keypoints $\{x_i^{\mathcal{S}},x_j^{\mathcal{T}}\}$ as an example, then the similarity features of this correspondence can be represented as $[s_{ij}^f,s_{ij}^b]$. The similarity features implicitly model bilateral consensus. If $x_j^\mathcal{T}$ is the most similar keypoint of $x_i^\mathcal{S}$ among all keypoints in $X^\mathcal{T}$, then $s_{ij}^f=1$. Then, if $x_i^\mathcal{S}$ is also the most similar keypoint in $X^{\mathcal{S}}$ of $x_j^\mathcal{T}$, $s_{ij}^b$ will also be equal to $1$, otherwise $s_{ij}^b<1$ because the best similarity score will not fall in $s_{ij}$ in this case. Thus, $s_{ij}^f$ and $s_{ij}^b$ will both be equal to $1$ only if the correspondence between $x_i^\mathcal{S}$ and $x_j^\mathcal{T}$ satisfies bilateral consensus.

\noindent \textbf{Neighborhood consensus:} In addition to bilateral consensus, neighborhood consensus is also important for good correspondence, which means that the neighboring keypoints of two corresponding keypoints should have similar features. To exploit neighborhood consensus, we propose an attention-based neighbor encoding module to gather the information of neighboring keypoints to generate neighbor-aware descriptors. Take a keypoint $x^\mathcal{S}$ in $X^\mathcal{S}$ as an example, we firstly perform $k$NN spatially to search $K$ neighboring keypoints in $X^\mathcal{S}$ to form a cluster. The features of the cluster consist of the descriptors of neighboring keypoints, relative coordinates and relative distances from neighboring to center keypoints. The cluster features are input into a Shared-MLP to generate a feature map. After that, a max-pool layer and a Softmax function are followed to predict attentive weights for each neighboring keypoint. The neighbor-aware descriptor $d_N^\mathcal{S}$ of $x^\mathcal{S}$ can be calculated as the weighted sum of the neighboring descriptors. Thus, the similarity of neighbor-aware descriptors can encode the similarity of neighboring keypoints. As shown in the bottom row of Fig.~\ref{fig:similarity}, using the neighbor-aware descriptors $D_N^{\mathcal{S}}$ and $D_N^{\mathcal{T}}$, we generate a neighbor-aware similarity matrix $S_N\in \mathbb{R}^{M\times M}$ through the similar method described before. 

Finally, the similarity features $F_S$ consist of two parts, namely $F_S^O$ and $F_S^N$, where $F_S^O$ denotes the similarity features from original similarity matrix $S$ and $F_S^N$ denotes that from the neighbor-aware similarity matrix $S_N$. Consequently, the introduction of similarity features $F_S$ is able to simultaneously incorporate bilateral consensus and neighborhood consensus into the registration pipeline implicitly.

\subsection{Fine registration}
After applying coarse registration in layer $l=3$, we obtain the coarse transformation $\mathbf{R}_3,\mathbf{t}_3$. Fine registration is applied in upper layers to reduce the registration error caused by the sparsity of the keypoints in deeper layers.  

Take the middle layer $l=2$ as an example, we firstly transform the source keypoints using the coarse transformation $\mathbf{R}_3,\mathbf{t}_3$. We assume that the coarse registration can provide a correct but not accurate enough estimation. Thus, the corresponding target keypoint $\tilde{x}^\mathcal{T}$ of a source keypoint $x^\mathcal{S}$ should be spatially close to $x^\mathcal{S}$ after the coarse transformation. Based on the above assumption, for a source keypoint $x^{\mathcal{S}}$, we perform $k$NN search locally in its spatial neighborhoods rather than in descriptor space to find $K$ candidate corresponding keypoints to construct a cluster. Different from coarse registration, the features of cluster in fine registration only include geometric features $F_G$ and descriptor features $F_D$. The similarity features are dropped here due to the computational complexity for a larger number of keypoints in upper layers. We then apply a similar correspondence network on the cluster to generate keypoints correspondences and confidence scores. Weighted Kabsch algorithm is followed to calculate the transformation $\Delta \mathbf{R}_2,\Delta \mathbf{t}_2$. Then the transformation $\mathbf{R}_2,\mathbf{t}_2$ after the fine registration in layer $l=2$ can be calculated as $\mathbf{R}_2 = \Delta\mathbf{R}_2\mathbf{R}_3, \mathbf{t}_2=\Delta\mathbf{R}_2\mathbf{t}_3 + \Delta\mathbf{t}_2$. Similarly, another fine registration is applied in the top layer $l=1$ based on the \textit{coarse transformation} $\mathbf{R}_2,\mathbf{t}_2$ to get the final registration result $\hat{\mathbf{R}},\hat{\mathbf{t}}$.

To summarize, the hierarchical structure leverages robust features in bottom layer and accurate position information in upper layers to achieve reliable and precise registration.

\subsection{Loss function}
The loss function $\mathcal{L}=\mathcal{L}_{trans}+\alpha \mathcal{L}_{rot}$, where $\mathcal{L}_{trans}$ and $\mathcal{L}_{rot}$ are translation and rotation loss, respectively. Given estimated and ground truth transformation $\hat{\mathbf{R}},\hat{\mathbf{t}}$ and $\mathbf{R},\mathbf{t}$, $\mathcal{L}_{trans}$ and $\mathcal{L}_{rot}$ can be calculated as
\begin{align}
    \mathcal{L}_{trans} &= \left\|\mathbf{t}-\hat{\mathbf{t}}\right\|_2
\label{eq:ltrans}
    \\
    \mathcal{L}_{rot} &= \left\|\hat{\mathbf{R}}^T\mathbf{R}-\mathbf{I}\right\|_2
    \label{eq:lrot}
\end{align}
where $\mathbf{I}$ denotes identity matrix.

\section{Experiments}

\subsection{Experiment settings}

\begin{figure*}
	\centering
	\subfigure{
		\includegraphics[width=0.31\textwidth]{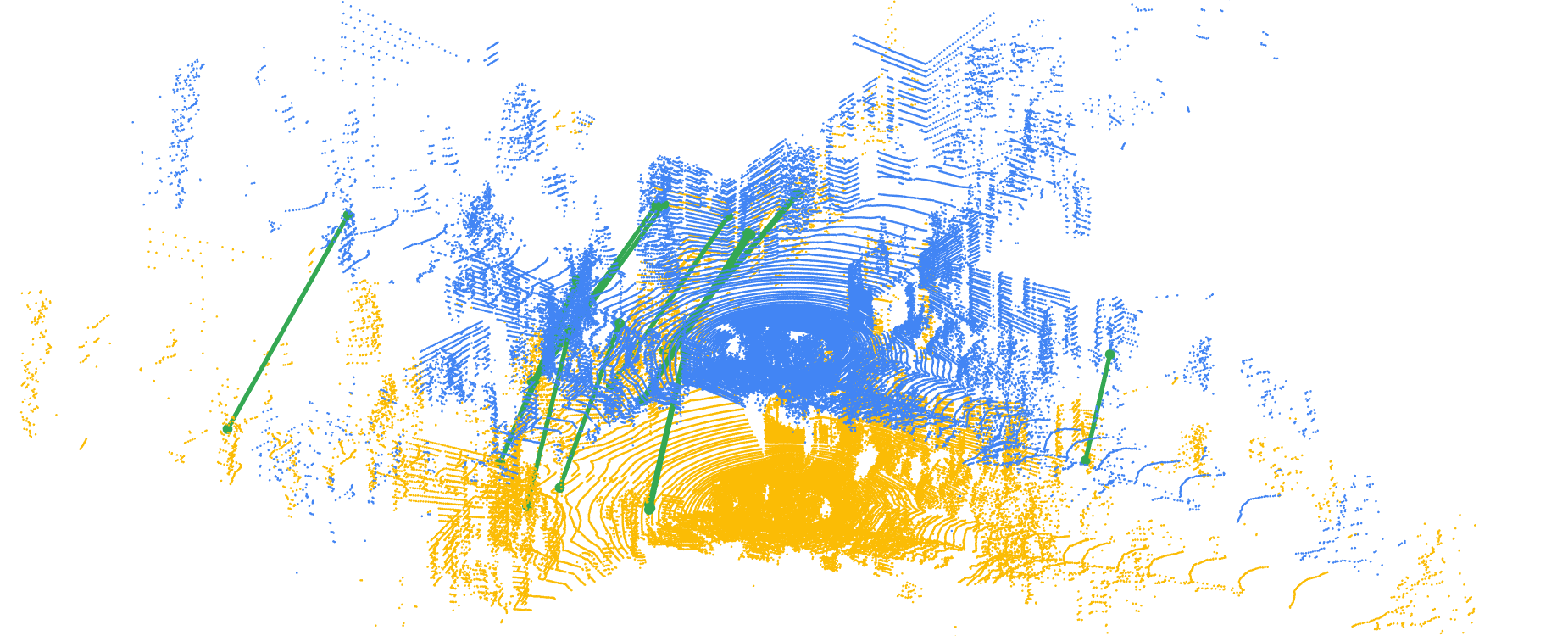}
	}
	\subfigure{
		\includegraphics[width=0.31\textwidth]{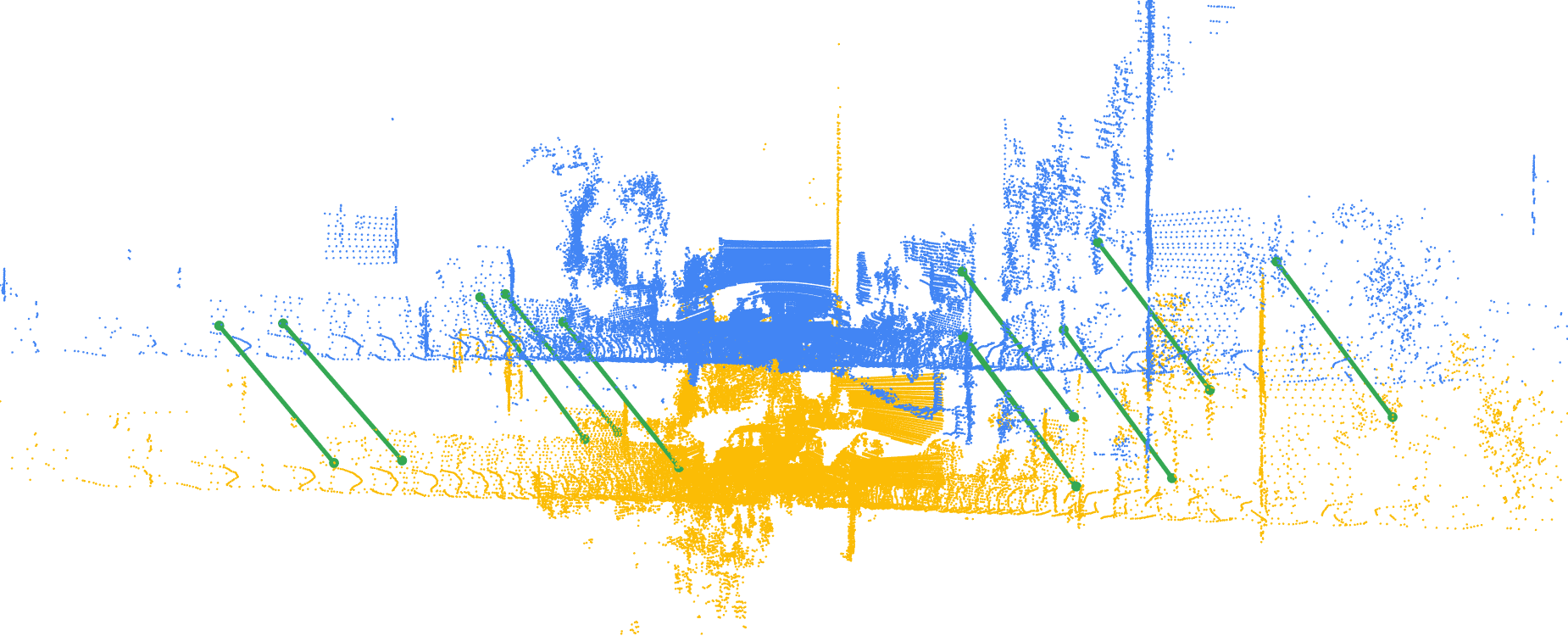}
	}
	\subfigure{
		\includegraphics[width=0.31\textwidth]{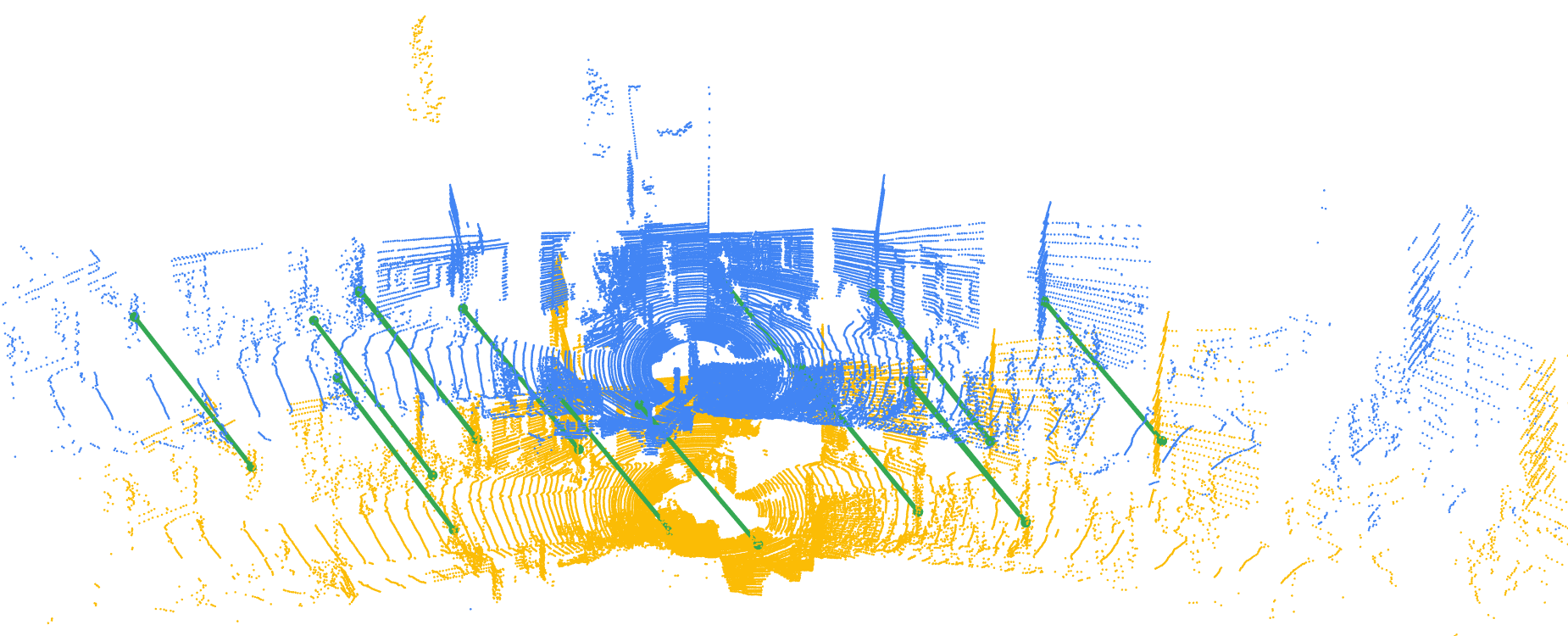}
	}\\
	\subfigure{
		\includegraphics[width=0.31\textwidth]{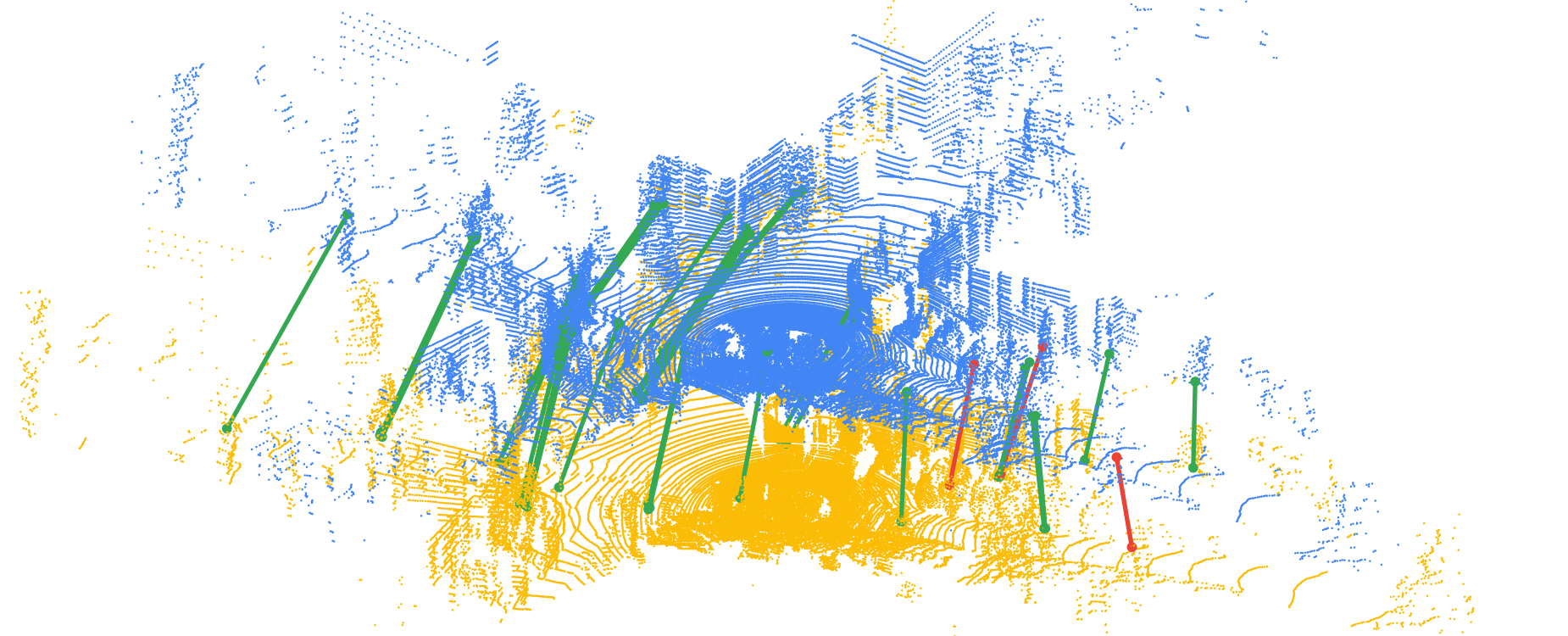}
	}
	\subfigure{
		\includegraphics[width=0.31\textwidth]{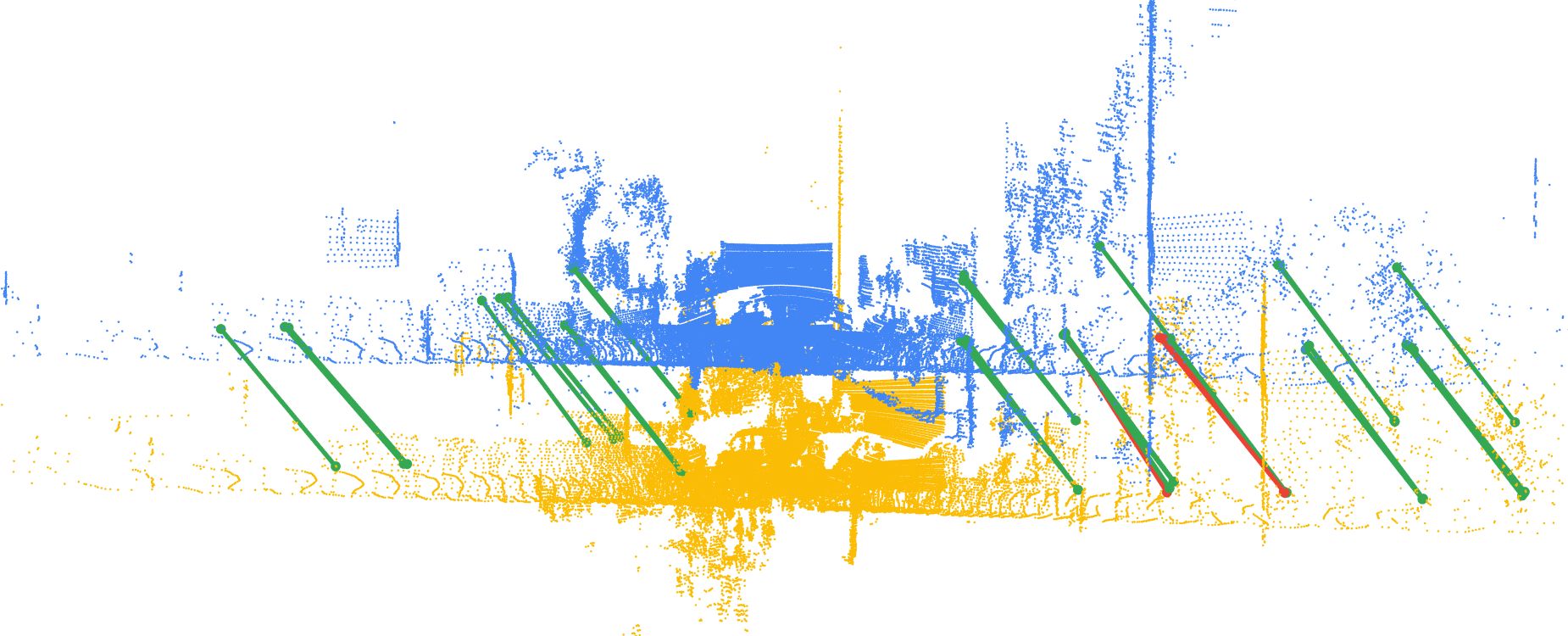}
	}
	\subfigure{
		\includegraphics[width=0.31\textwidth]{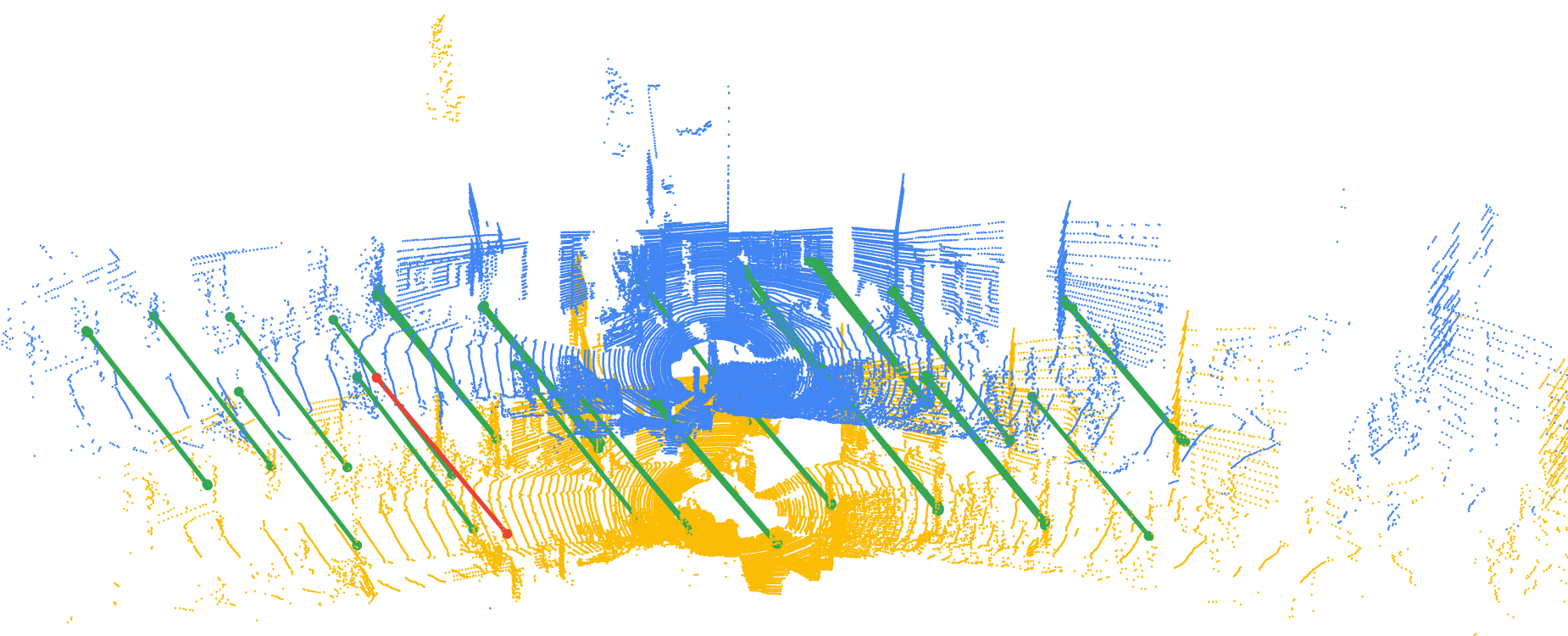}
	}\\
	\subfigure{
		\includegraphics[width=0.31\textwidth]{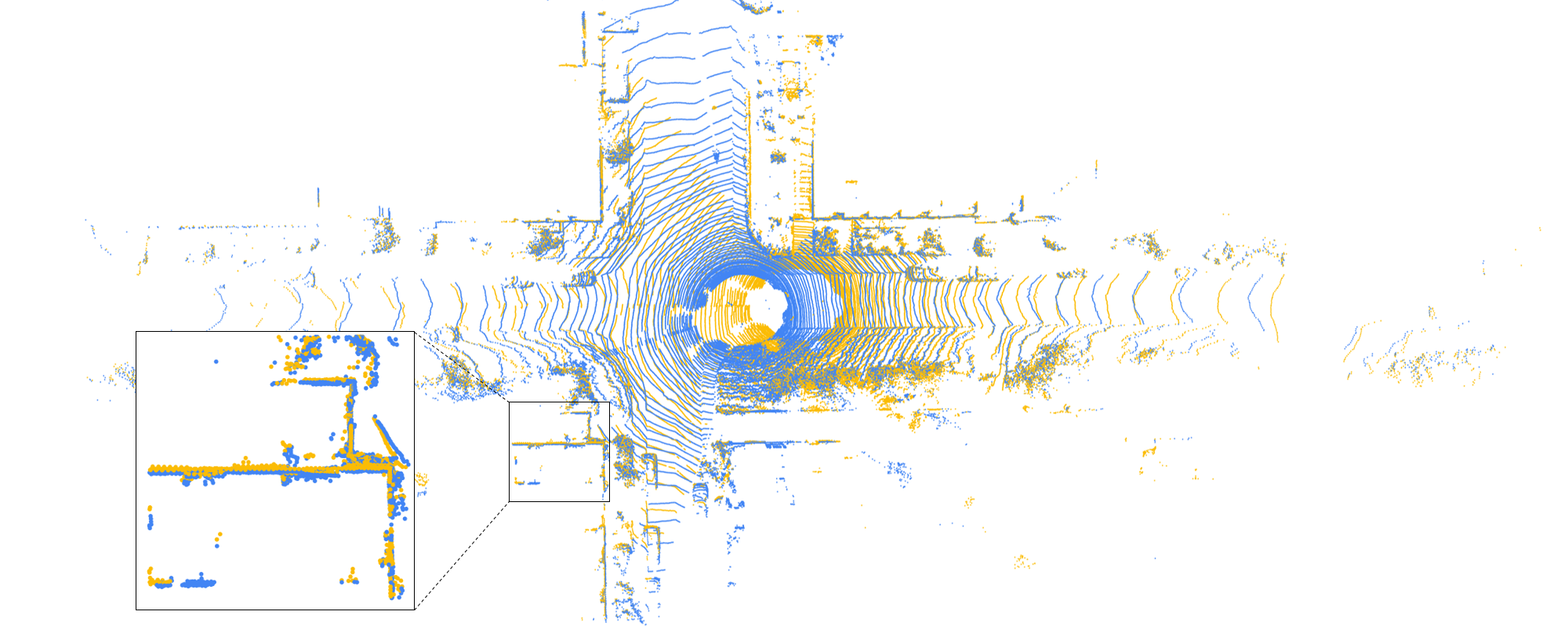}
	}
	\subfigure{
        \includegraphics[width=0.31\textwidth]{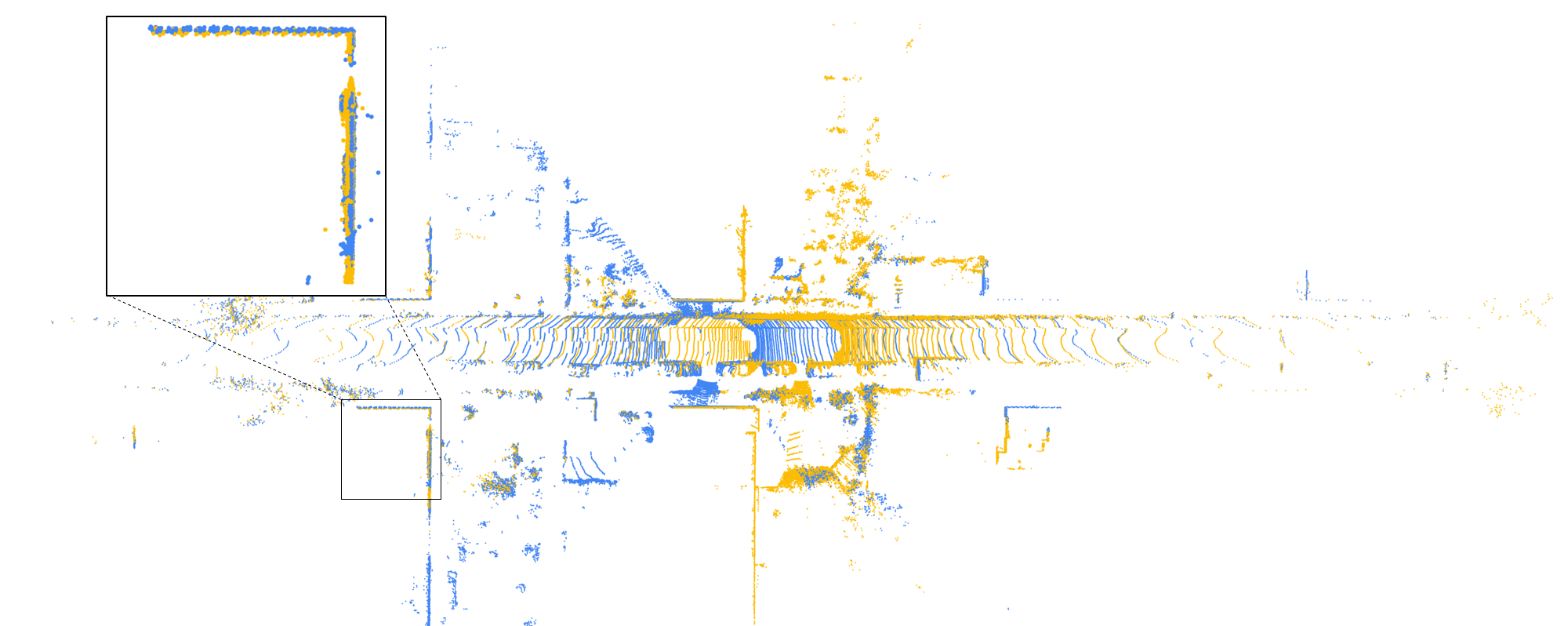}
	}
	\subfigure{
        \includegraphics[width=0.31\textwidth]{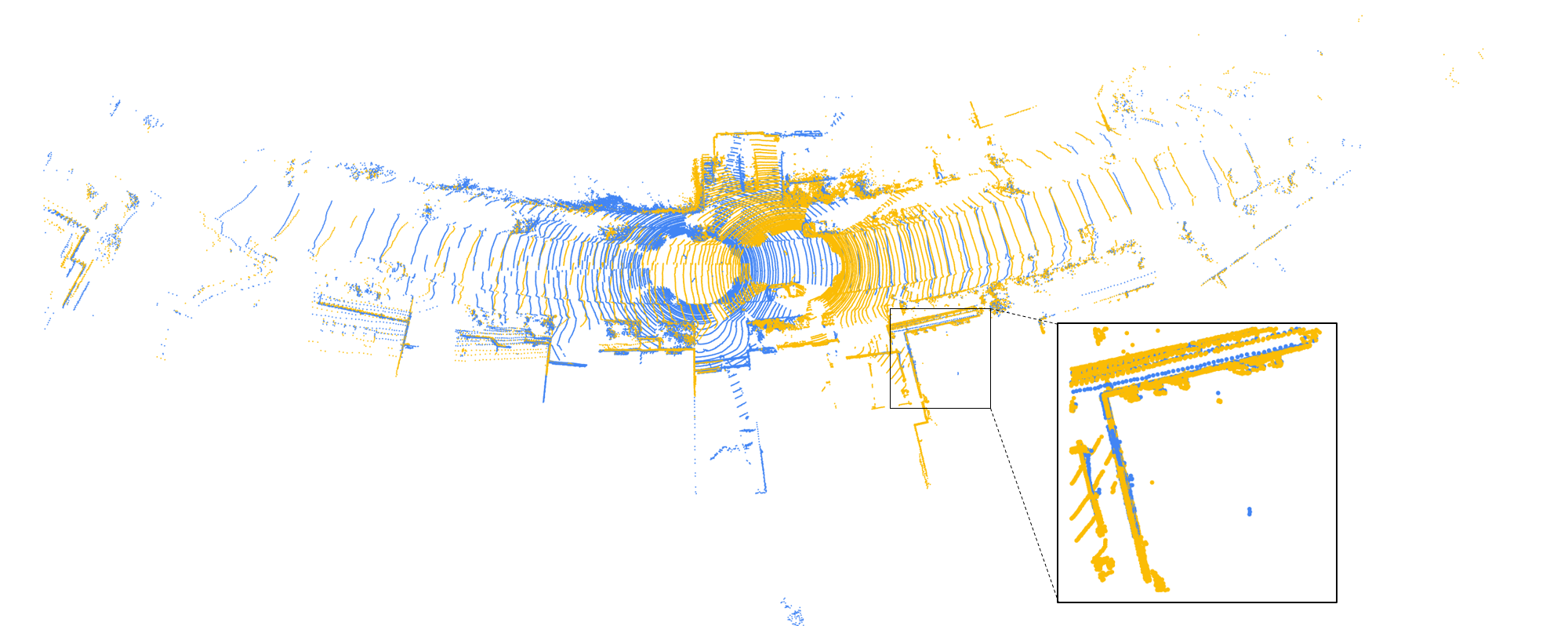}
	}\\
	\caption{Qualitative visualization of the proposed point cloud registration method. We display 3 samples of point cloud registration here. The first row displays the correspondences between source and target keypoints in coarse registration with confidence score $\tilde{c}>0.005$ and the second row displays the correspondences with confidence score $\tilde{c}>0.0005$. The green lines and red lines represent inlier and outlier correspondences, respectively. The bottom row shows the aligned two point clouds and we zoom in an area for better visualization.}
	\label{fig:registration}
\end{figure*}

\noindent \textbf{Datasets:} We perform extensive experiments on two large-scale outdoor LiDAR point cloud datasets, namely KITTI odometry dataset \cite{geiger2012we} and NuScenes dataset \cite{caesar2020nuscenes}. KITTI dataset consists of 11 sequences (00 to 10) with ground truth vehicle poses and we use Sequence 00 to 05 for training, 06 to 07 for validation and 08 to 10 for testing. We use the current frame with the 10th frame after that to form a pair of point clouds. To reduce the noise of ground truth vehicle poses, we perform Iterative Closest Point (ICP) algorithm in Open3D library \cite{Zhou2018} to refine the noisy relative transformation between two point clouds. NuScenes dataset includes 1000 scenes, among which 850 scenes are used for training and validation and 150 scenes for testing. We use the first 700 scenes in the 850 scenes to train the network and the other 150 scenes for validation. NuScenes dataset only provides the ground truth poses of the given samples and the time interval between two consecutive point cloud samples is about 0.5s. We use the current point cloud sample with the second sample after it as a pair of point clouds.

\noindent \textbf{Implementation details:} In the pre-processing, we firstly voxelize the input point clouds and the voxel size is set to 0.3m. After that, we randomly sample 16384 points from the point clouds in KITTI dataset and 8192 points in NuScenes dataset. The network is implemented using PyTorch \cite{pytorch} and we use Adam \cite{adam} as the optimizer. The learning rate is initially set to 0.001 and decreases by 50\% every 10 epochs. The hyperparameter $\alpha$ in the loss function $\mathcal{L}$ is set to 1.8 for KITTI dataset and 2.0 for NuScenes dataset. When training the network, we firstly pre-train the feature extraction module and then train the whole network based on the pre-trained features. The whole network is trained on an NVIDIA RTX 3090 GPU. The details of pre-training and the network architecture are described in the supplementary material.  

\noindent \textbf{Baseline methods:}
We compare the performance of the proposed HRegNet with both classical and learning-based methods. All of the methods are tested on an Intel i9-10920X CPU and an NVIDIA RTX 3090 GPU.

\textit{Classical methods}: We evaluate the performance of point to point ICP (ICP (P2Point)), point to plane ICP (ICP (P2Plane)) \cite{icp}, RANSAC \cite{fischler1981random}, and Fast Global Registration (FGR) \cite{zhou2016fast}. All of the classical methods are implemented using Open3D library \cite{Zhou2018}. For RANSAC and FGR, we extract Fast Point Feature Histograms (FPFH) \cite{rusu2009fast} from 0.3m-voxel-downsampled point clouds. The maximum iteration number of RANSAC is set to 2e6\footnote{We have tried more iterations, however, the accuracy will not be obviously improved while the computational time will increase significantly.}.

\textit{Learning-based methods}: We choose 4 representative learning-based methods to compare with the proposed HRegNet\footnote{We also try to compare our method with DeepVCP\cite{lu2019deepvcp}, however, the source code has not been released by the author and the self-implemented version does not provide reasonable results.}. (1) Deep Closest Point (DCP) \cite{wang2019deep}: DCP is a pioneering work for learning-based point cloud registration. For the pre-processing of point clouds, 4096 points are randomly sampled from 0.3m-voxel-downsampled point clouds for both datasets. (2) IDAM \cite{idam}: IDAM is one of the state-of-the-art object-level point cloud registration methods. The pre-processing is the same as that for DCP. (3) Feature-metric Registration (FMR) \cite{huang2020feature}: FMR has been evaluated for both object-level and indoor point cloud registration. The pre-processing of point clouds is the same as that in our methods. (4) Deep Global Registration (DGR) \cite{choy2020deep}: DGR achieves state-of-the-art performance in indoor point cloud registration. The point clouds are voxelized with 0.3m voxel size. All the learning-based baseline methods are retrained on both datasets for better performance. 

\subsection{Evaluation}

\begin{figure*}
	\centering
	\subfigure{
		\includegraphics[width=0.8\textwidth]{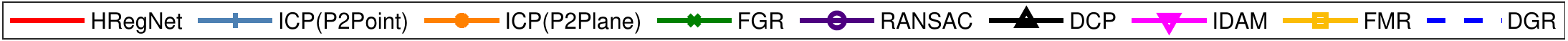}
	}\\
	\begin{minipage}[t]{0.49\linewidth}
		\includegraphics[width=0.49\textwidth]{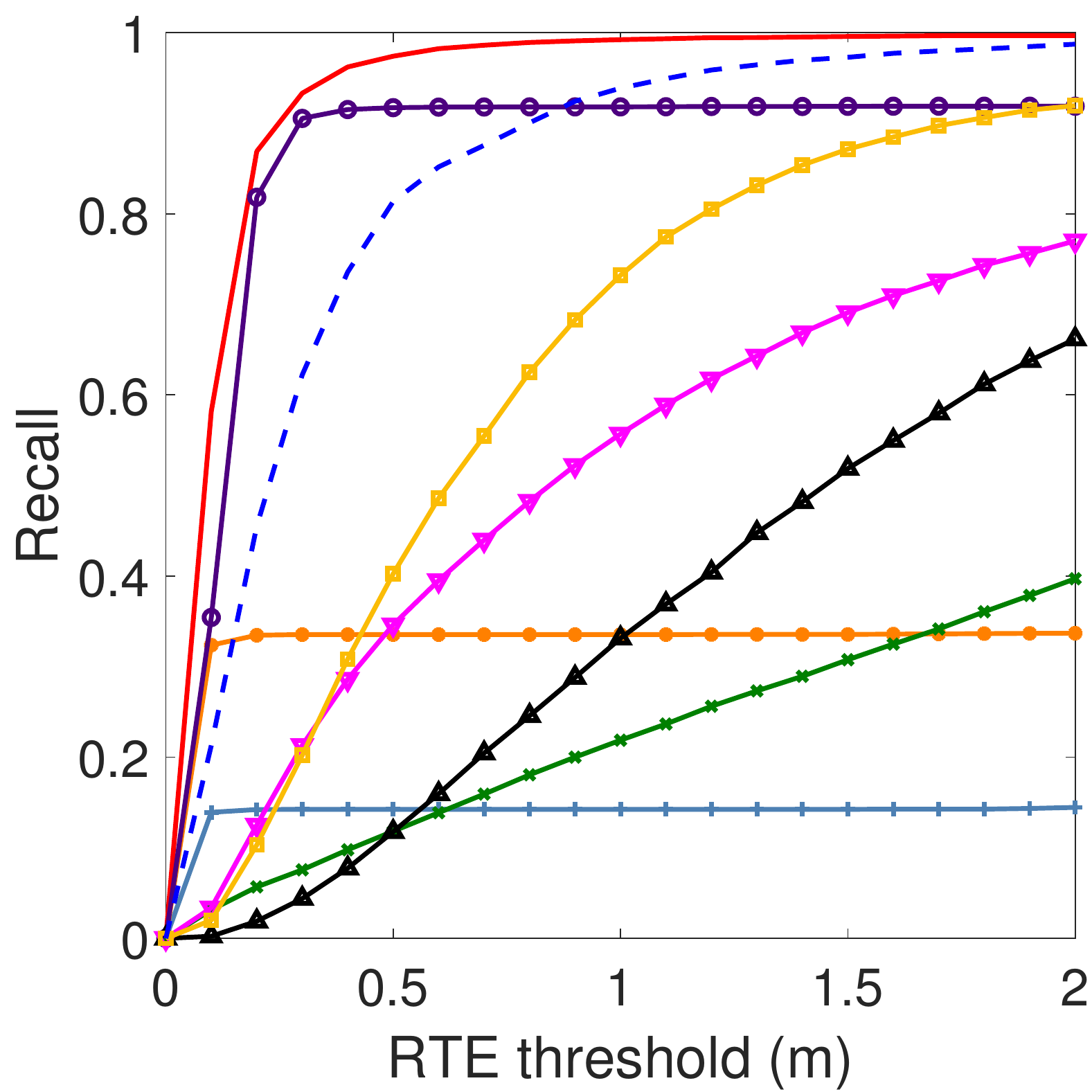}
		\includegraphics[width=0.49\textwidth]{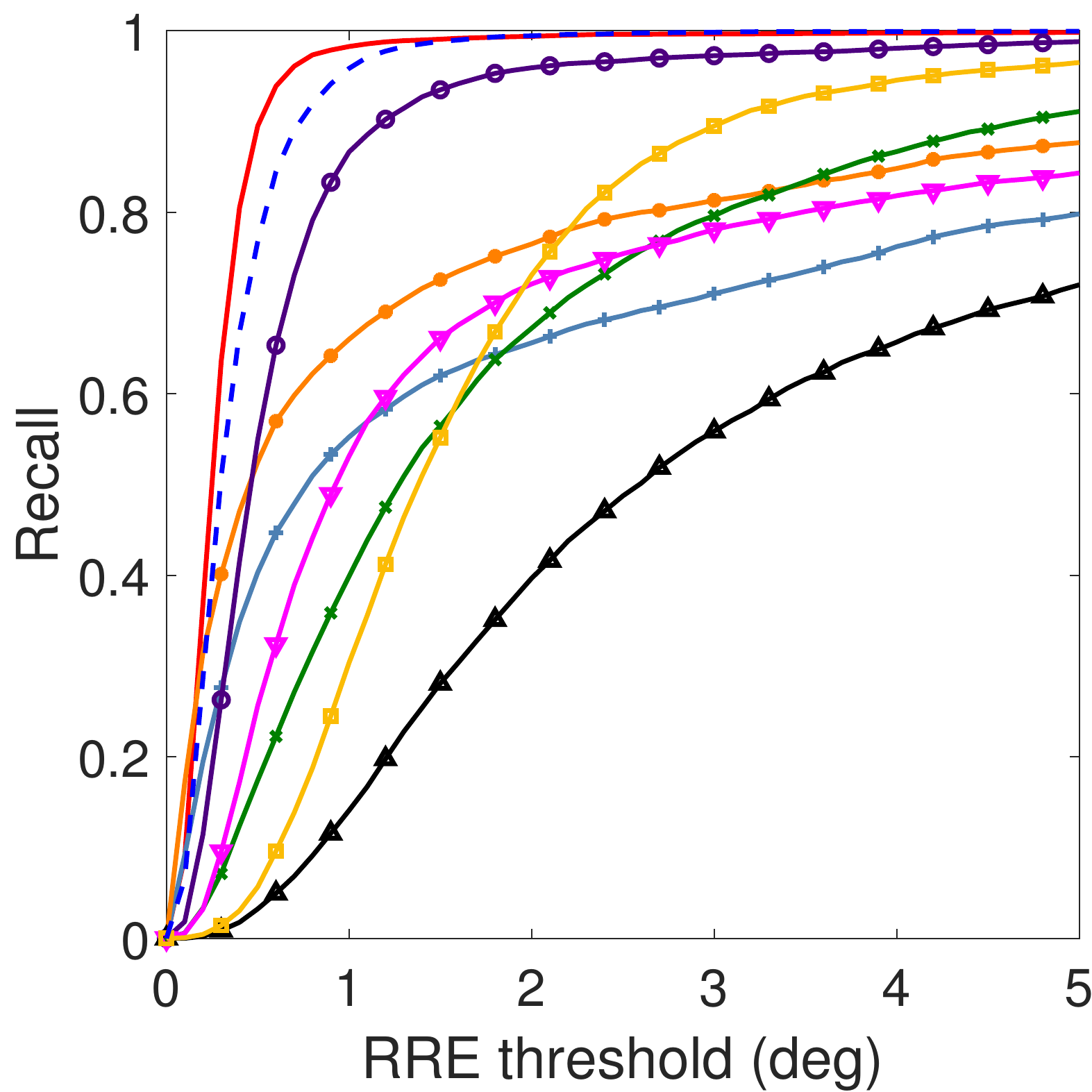}
		\centerline{\footnotesize{(a) KITTI dataset}}
	\end{minipage}
	\begin{minipage}[t]{0.49\linewidth}
		 \includegraphics[width=0.49\textwidth]{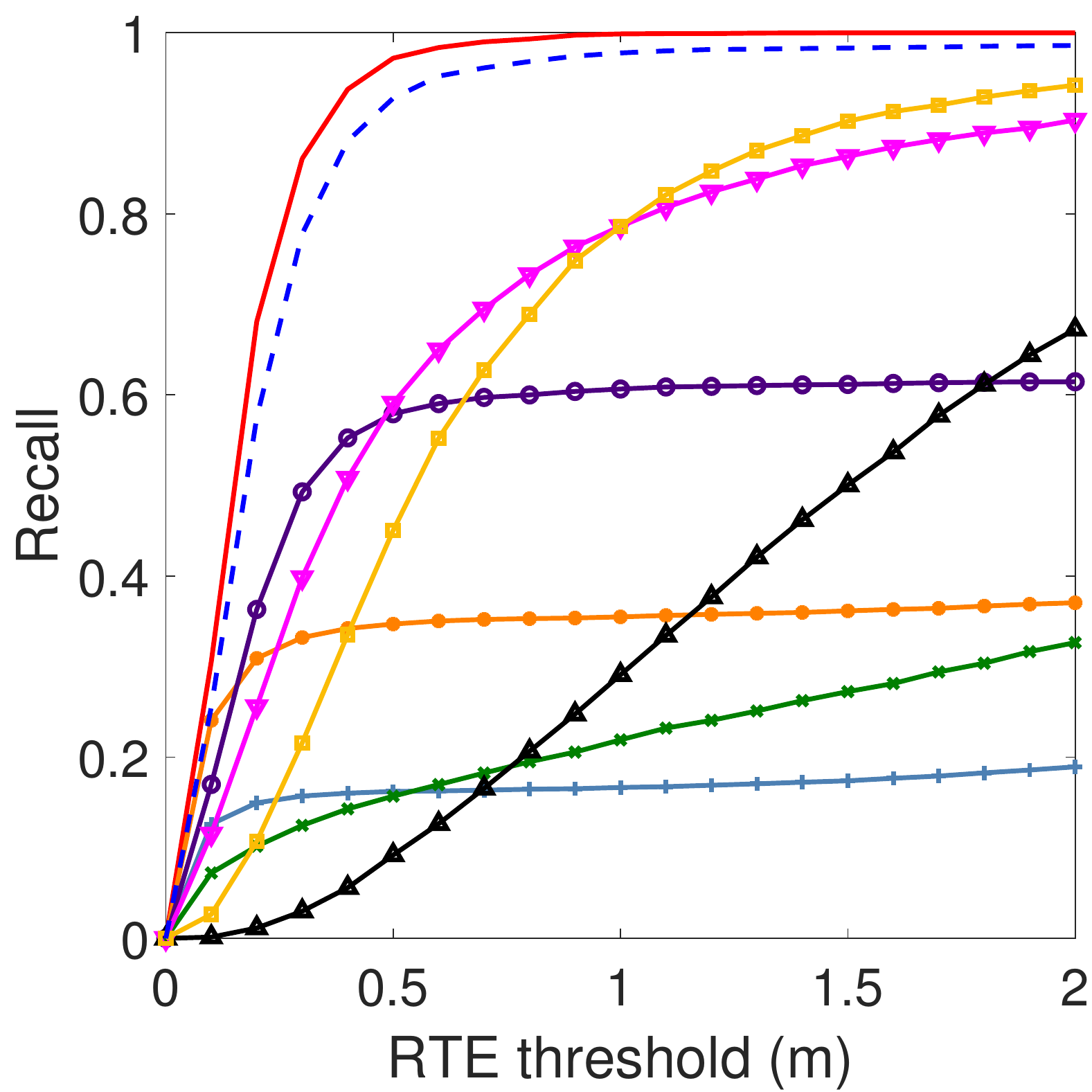}
		\includegraphics[width=0.49\textwidth]{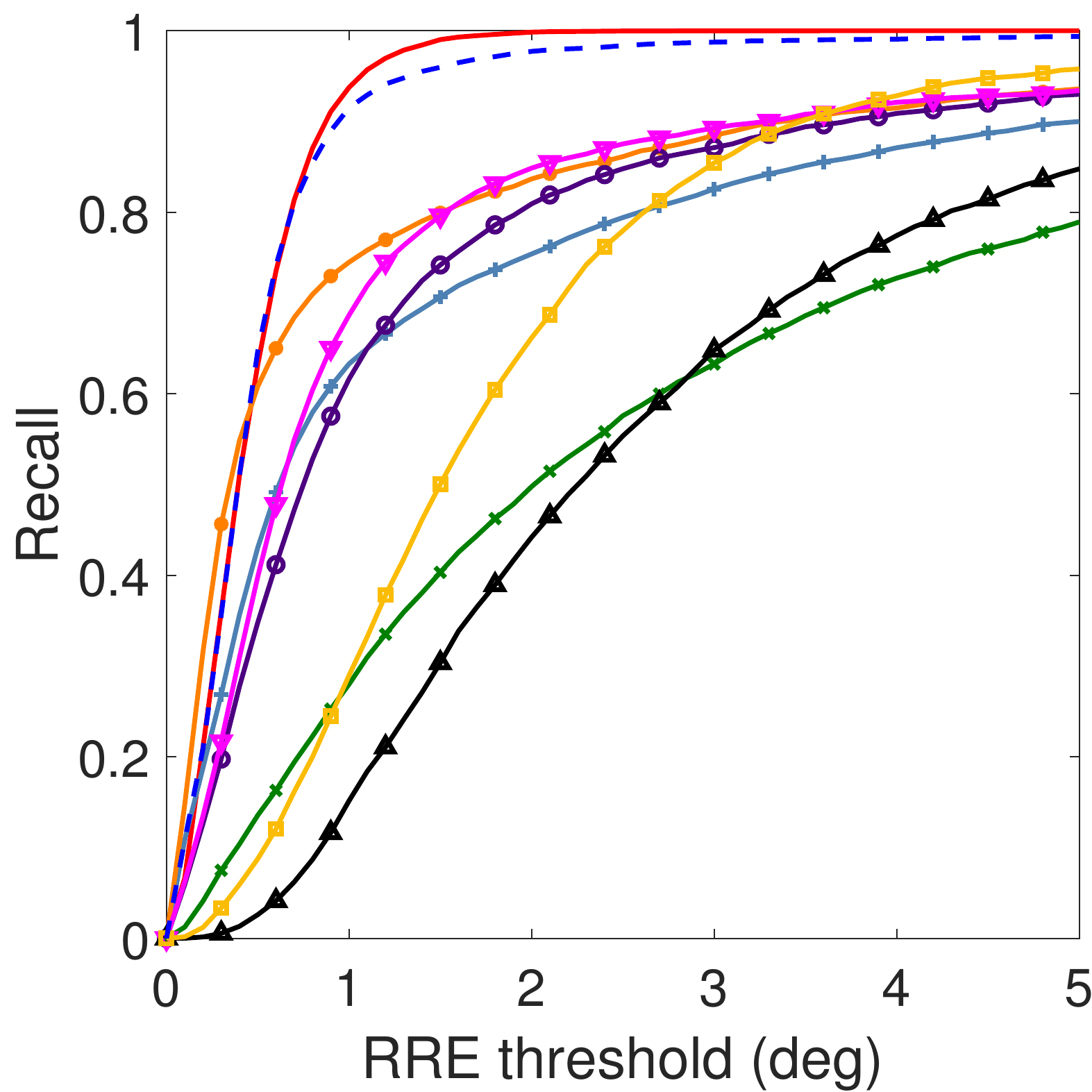}
		\centerline{\footnotesize{(b) NuScenes dataset}}
	\end{minipage}

    \caption{Registration recall with different RRE and RTE thresholds on KITTI dataset and NuScenes dataset. }
	\label{fig:recall}
	\vspace{-4mm}
\end{figure*}

\paragraph{Qualitative visualization:} 
We display several qualitative samples of point cloud registration in Fig.~\ref{fig:registration}. Corresponding keypoints in coarse registration with confidence scores $\tilde{c} > 0.005$ and $\tilde{c}>0.0005$ are displayed in the first and second row respectively. Two corresponding keypoints are considered as an inlier if the relative position error (after applying the ground truth relative transformation) less than a distance threshold $\epsilon_d=1$m. The green and red lines represent inlier and outlier correspondences, respectively. According to the results, the correspondences with larger confidence score ($\tilde{c} > 0.005$) are basically all inliers and several mismatches start to appear when reducing the threshold of $\tilde{c}$ to 0.0005. The qualitative results show that the correspondence network can generate accurate and correct correspondence of keypoints and the predicted confidence score can effectively reject unreliable correspondences. The third row of Fig.~\ref{fig:registration} displays the two aligned point clouds, which demonstrates that the network can precisely predict the transformation. More qualitative results are displayed in our supplementary material.

\begin{table*}
	\centering
	\caption{Registration performance on KITTI dataset and NuScenes dataset.}
	\label{tab:reg}
	\footnotesize
	\begin{tabular}{l|llll|llll}
		\toprule
		\multirow{2}{*}{Methods}&\multicolumn{4}{|c|}{KITTI dataset}&\multicolumn{4}{c}{NuScenes dataset}\\
		\cmidrule(r){2-5}
		\cmidrule(r){6-9}
		~ & RTE (m) & RRE (deg) & Recall & Time (ms) & RTE (m) & RRE (deg) & Recall & Time (ms)\\
		\midrule
		ICP (P2Point) \cite{icp} & $0.04\pm 0.05$ & $0.11\pm 0.09$ & 14.3\% & 472.2 & $0.25\pm 0.51$ & $0.25\pm 0.50$ & 18.8\% & 82.0 \\
		ICP (P2Plane) \cite{icp} & $0.04\pm 0.04$ & $0.14\pm 0.15$ & 33.5\% & 461.7 & $0.15\pm 0.30$ & $0.21\pm 0.31$ & 36.8\% & 44.5 \\
		FGR \cite{zhou2016fast} & $0.93\pm 0.59$ & $0.96\pm 0.81$ & 39.4\% & 506.1 & $0.71\pm 0.62$ & $1.01\pm 0.92$ & 32.2\% & 284.6 \\
		RANSAC \cite{fischler1981random} & $0.13\pm 0.07$ & $0.54\pm 0.40$ & 91.9\% & 549.6 & $0.21\pm 0.19$ & $0.74\pm 0.70$ & 60.9\% & 268.2 \\
		
		\midrule
		DCP \cite{wang2019deep} & $1.03\pm 0.51$ & $2.07\pm 1.19$ & 47.3\% & 46.4 & $1.09\pm 0.49$ & $2.07\pm 1.14$ & 58.6\% & 45.5\\
		IDAM \cite{idam} & $0.66\pm 0.48$ & $1.06\pm 0.94$ & 70.9\% & 33.4 & $0.47\pm 0.41$ & $0.79\pm 0.78$ & 88.0\% & 32.6 \\
		FMR \cite{huang2020feature} & $0.66\pm 0.42$ & $1.49\pm 0.85$ & 90.6\% & 85.5 & $0.60\pm 0.39$ & $1.61\pm 0.97$ & 92.1\% & 61.1 \\ 
		DGR \cite{choy2020deep} & $0.32\pm 0.32$ & $0.37\pm 0.30$ & 98.7\% & 1496.6 & $0.21\pm 0.18$ & $0.48\pm 0.43$ & 98.4\% & 523.0\\
		\midrule
		HRegNet & $\mathbf{0.12\pm 0.13}$ & $\mathbf{0.29\pm 0.25}$ & \textbf{99.7}\% & 106.2 & $\mathbf{0.18\pm 0.14}$ & $\mathbf{0.45\pm 0.30}$ & \textbf{99.9}\% & 87.3 \\
		\bottomrule
	\end{tabular}
	\vspace{-3mm}
\end{table*}

\paragraph{Quantitative evaluation:} We adopt relative translation error (RTE) and relative rotation error (RRE) to evaluate the registration performance. RTE can be calculated as Eq.~\ref{eq:ltrans} and RRE can be represented as ${\rm arccos}({{\rm Tr}(\hat{\mathbf{R}}^T\mathbf{R}-1)}/{2})$, where $\hat{\mathbf{R}}$ and $\mathbf{R}$ are the estimated and ground truth rotation matrix. Registration recall is defined as the ratio of successful registration. A registration is considered as successful when the RTE and RRE are within the thresholds $\epsilon_{trans}$ and $\epsilon_{rot}$. We display the registration recall with different RTE and RRE thresholds on two datasets in Fig.~\ref{fig:recall}. According to the results, the proposed HRegNet outperforms all baseline methods by an obvious margin on both two datasets. Besides, for a more detailed comparison of the registration performance, we calculate the average RRE and RTE and display the results in Table~\ref{tab:reg}. Noting that a part of failed registrations can result in dramatically large RRE and RTE, which can cause unreliable error metrics. Thus, the average RTE and RRE are only calculated for successful registrations and the thresholds are set as $\epsilon_{trans}=2$m and $\epsilon_{rot}=5$deg. The registration recall at the given threshold is also displayed in Table~\ref{tab:reg}.

According to the results, ICP algorithms (for both ICP (P2Point) and ICP (P2Plane)) fail to generate reasonable relative transformation in most cases due to the lack of precise initial transformation between two point clouds. FGR performs slightly better than ICP, however, the registration recall is still below 50\%, which is unacceptable in applications. RANSAC achieves the best performance among the classical methods thanks to the powerful outlier rejection mechanism, however, the iterative paradigm can also result in poor efficiency. The average RTE of RANSAC is similar to ours method, however, it is due to a number of mismatches are omitted in the calculation and the registration recall of RANSAC is obviously lower than the proposed method according to Fig.~\ref{fig:recall}. Moreover, the runtime of our method is almost 1/5 of RANSAC on KITTI dataset.

As for the learning-based methods, the recall of DCP on KITTI and NuScenes dataset are both less than 60\% and the average RTE and RRE are also quite large. IDAM performs better than DCP, however, the recall is still only about 70\% on KITTI dataset and the RTE and RRE are much higher than the proposed method, which indicates the poor applicability of the object-level point cloud registration methods to complex large-scale LiDAR point clouds. FMR achieves a slightly faster speed than our method, however, the registration error is much higher than ours. For example, the RTE of FMR on KITTI dataset is more than 5 times of our method. DGR achieves the best registration performance among all the learning-based baseline methods. However, the 6D convolutional network-based outlier rejection method is time-consuming and the voxel-based representation of point clouds limits the precision of the registration. The RTE of our method is almost 1/3 of that of DGR on KITTI dataset. Moreover, our method achieves almost $15\times$ faster speed than DGR on KITTI dataset. 

Overall, extensive experiments demonstrate that the proposed HRegNet achieves state-of-the-art performance in terms of both accuracy and efficiency.

\subsection{Ablation study}
We perform abundant ablation studies on KITTI dataset to demonstrate the effectiveness of the hierarchical structure and the introduction of the similarity features.

\begin{figure}
	\centering

	\subfigure{
		\includegraphics[width=0.30\textwidth]{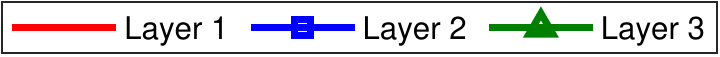}

	}\\
	\begin{minipage}[t]{0.99\linewidth}
		\includegraphics[width=0.49\textwidth]{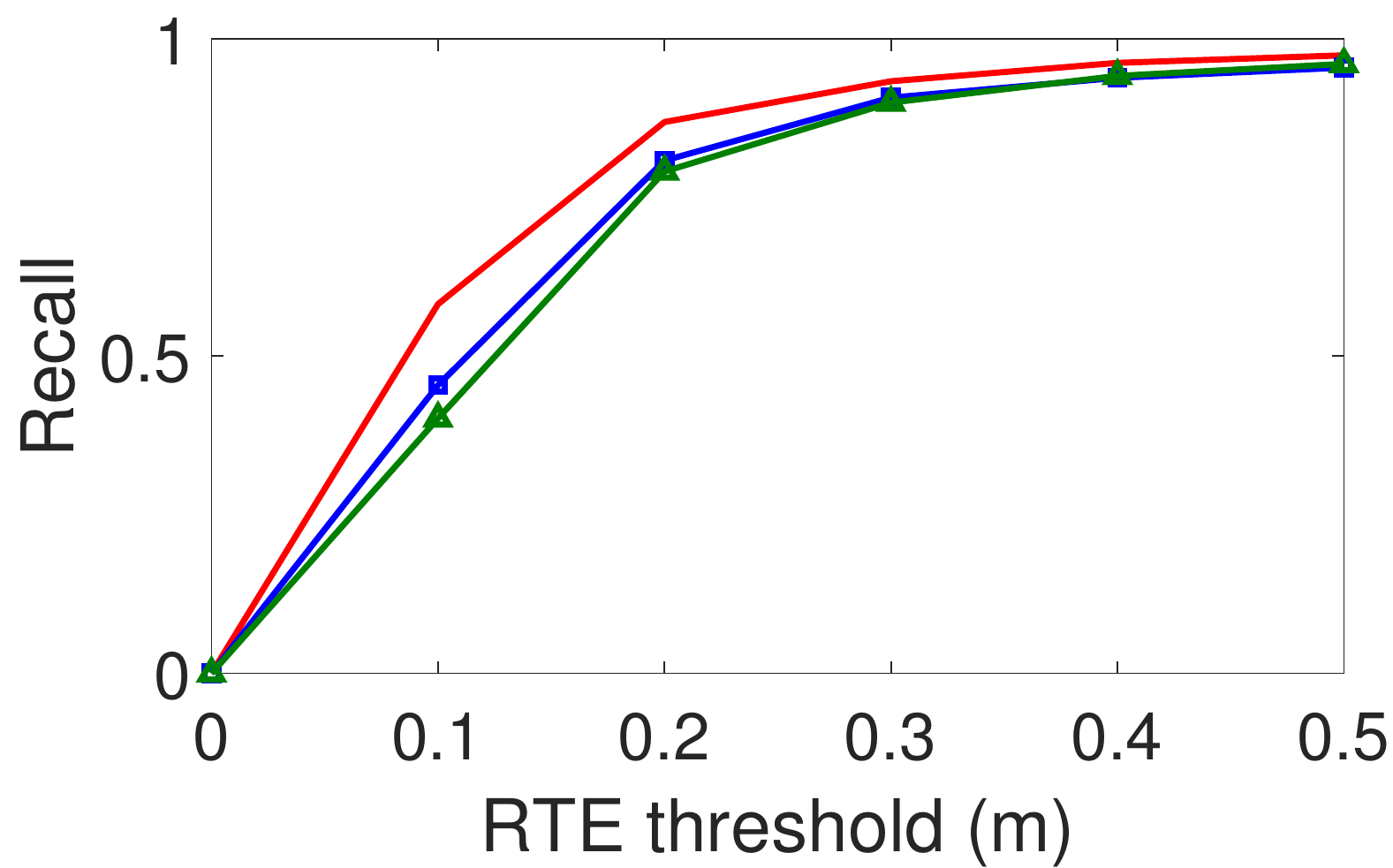}
		\includegraphics[width=0.49\textwidth]{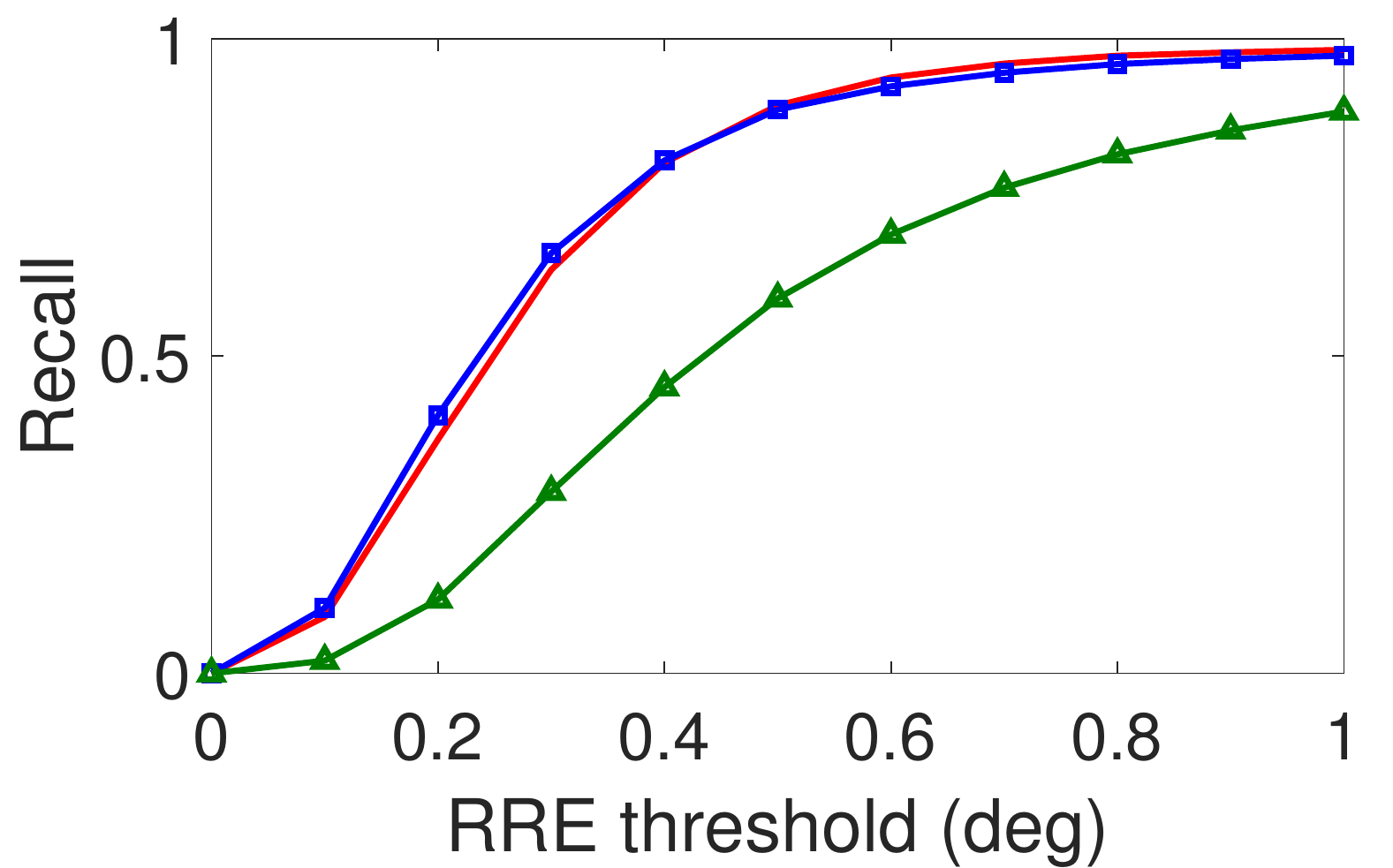}
	\end{minipage}\\
	
	\caption{Registration recall of different output layers on KITTI dataset. We set the range of RTE threshold as from 0 to 0.5m and RRE threshold as from 0 to 1deg for better visualization. Layer 1: top layer; Layer 2: middle layer; Layer 3: bottom layer.}
	\label{fig:hier}
	\vspace{-5mm}
\end{figure}

\noindent \textbf{Hierarchical structure:} To validate the effectiveness of the hierarchical structure, we use the output transformation $\mathbf{R},\mathbf{t}$ from layer 3 to layer 1 as the final estimation respectively to evaluate the performance. The network with different output layers is trained separately using the same hyperparameters. The registration recall with different output layers is displayed in Fig.~\ref{fig:hier}. The detailed average RRE and RTE is shown in Table~\ref{tab:ablation} and the calculation settings are the same as that in Table~\ref{tab:reg}. According to the results, the average RTE and RRE are gradually reduced with the layer-by-layer refinement. The results in layer 2 achieve much lower rotation error than layer 3. And the translation accuracy in layer 1 (\emph{i.e.}, the full model) is also obviously improved compared to layer 2, which demonstrates the validity of hierarchical refinement strategy. Noting that the registration recall with different RRE thresholds of layer 1 is almost the same as layer 2 and we found that further increasing the number of layers will not result in significant improvements in registration performance, however, will deteriorate the efficiency of the network. Considering the trade-off between accuracy and efficiency, we choose the 3-layer implementation.

\begin{figure}
	\centering
	\subfigure{
		\includegraphics[width=0.45\textwidth]{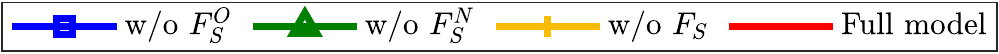}
	}\\
	\begin{minipage}[t]{0.99\linewidth}
		\includegraphics[width=0.49\textwidth]{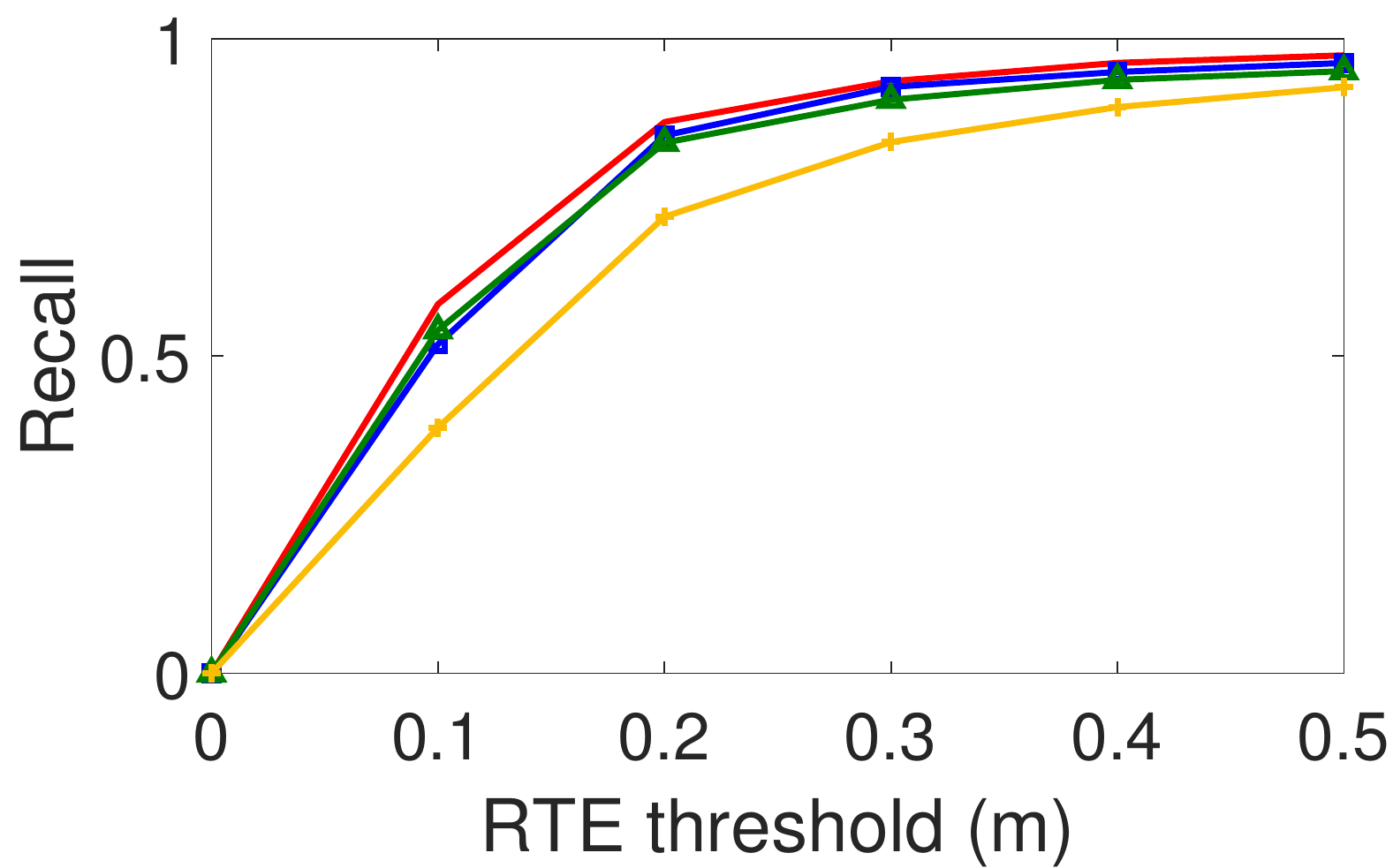}
		\includegraphics[width=0.49\textwidth]{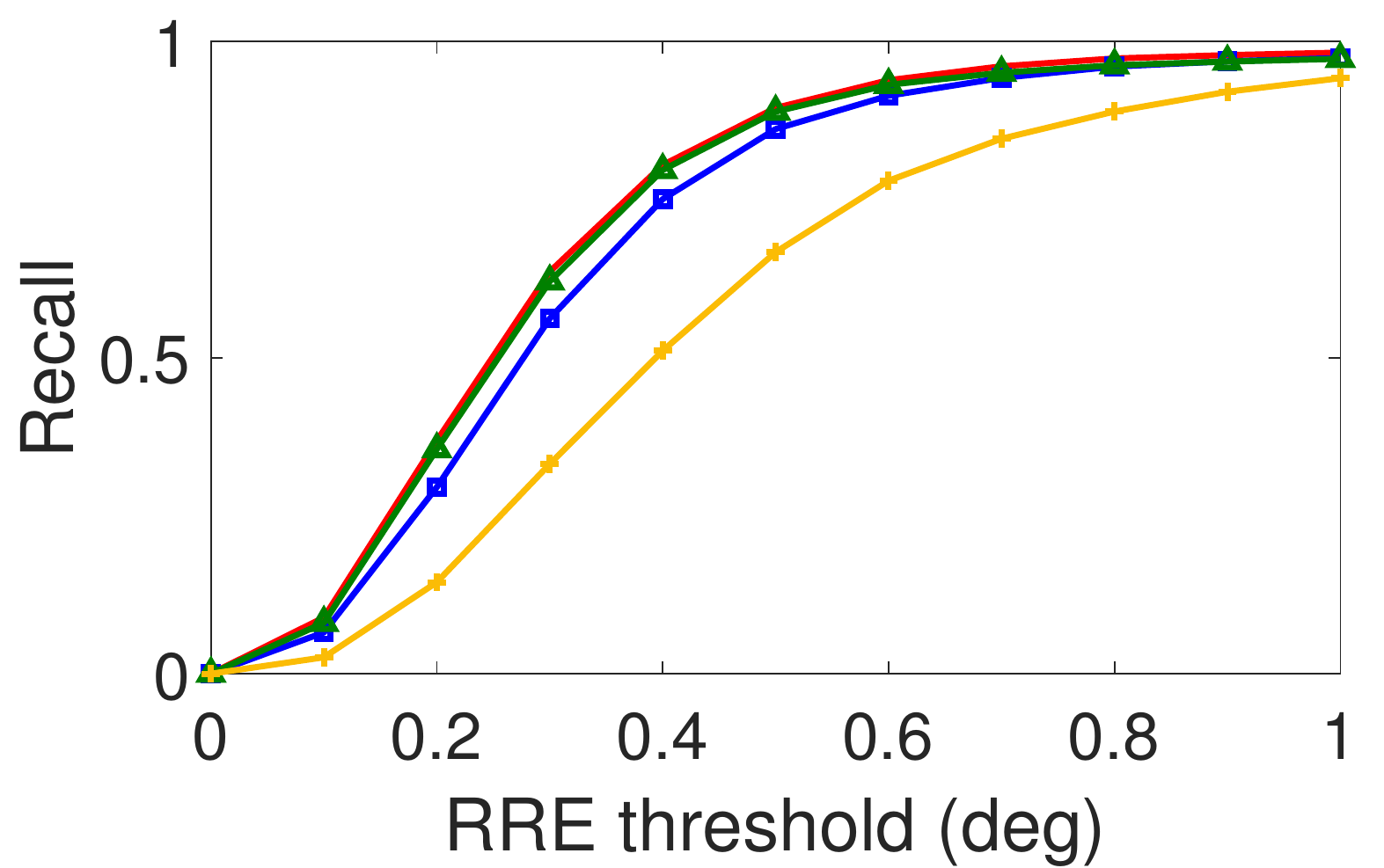}
	\end{minipage}\\
	
	\caption{Registration recall with and without (w/o) similarity features on KITTI dataset. $F_S^O$: original similarity features. $F_S^N$: neighbor-aware similarity features. $F_S$: both similarity features.}
	\label{fig:sim}
\end{figure}

\noindent \textbf{Similarity features:}
As we described before, the similarity features $F_S$ consist of two parts, namely the original similarity features $F_S^O$ and neighbor-aware similarity features $F_S^N$. To analysis the impact of the two parts on the performance, we drop $F_S^O$ and $F_S^N$ separately and retrain the network. The registration recall and the average RRE and RTE of the full model and the model without $F_S^O$, $F_S^N$ and $F_S$ are displayed in Fig.~\ref{fig:sim} and Table~\ref{tab:ablation}. According to the results, the registration recall without both similarity features $F_S$ is inferior to the other cases by a significant margin, which demonstrates the importance of the bilateral consensus. The neighbor-aware similarity features $F_S^N$ incorporate the information of neighboring keypoints into consideration, however, may also lead to the neglect of the own unique features of the keypoint. Thus, the original and neighbor-aware similarity features are complementary to each other and the combination of the two (\emph{i.e.}, the full model) outperforms other cases. Overall, the results demonstrate that the introduction of the similarity features significantly improves the performance.

\begin{table}
	\centering
	\caption{Ablation studies on KITTI dataset.}
	\label{tab:ablation}
	\footnotesize
	\begin{tabular}{lllll}
		\toprule
		Model & RTE (m) & RRE (deg) & Recall & Time (ms) \\
		\midrule
		Full & $\mathbf{0.12\pm0.13}$ & $\mathbf{0.29\pm 0.25}$ & \textbf{99.7}\% & 106.2 \\
		\midrule
		Layer 2 & $0.15\pm 0.18$ & $0.29 \pm 0.27$ & 99.2\% & 101.4 \\
		Layer 3 & $0.16\pm 0.18$ & $0.55 \pm 0.45$ & 99.7\% & 96.9 \\
		\midrule
		w/o $F_S^O$ & $0.15\pm 0.19$ & $0.31\pm 0.30$ & 99.1\% & 98.6 \\
		w/o $F_S^N$ & $0.14\pm 0.17$ & $0.33\pm 0.29$ & 99.4\% & 96.4 \\
		w/o $F_S$ & $0.19\pm 0.22$ & $0.46\pm 0.36$ & 98.7\% & 88.0 \\
		\bottomrule
	\end{tabular}
	\vspace{-5mm}
\end{table}

\section{Conclusion}
In this paper, we provide an efficient hierarchical network for large-scale outdoor LiDAR point cloud registration. The hierarchical paradigm leverages different characteristics of keypoints and descriptors in deeper and shallower layers by introducing coarse registration and fine registration in different layers. To construct reliable correspondences between keypoints, we propose a correspondence network to generate corresponding keypoints. Moreover, novel similarity features are designed to effectively incorporate bilateral consensus and neighborhood consensus into the registration pipeline. Abundant ablation studies demonstrate the effectiveness of the hierarchical paradigm and the introduction of similarity features. Besides, the network is also highly efficient since we only use a small number of keypoints for registration. Extensive experiments on two large-scale LiDAR point cloud datasets demonstrate the high accuracy and efficiency of the proposed HRegNet. 

\paragraph{Acknowledgments:}This work is funded by the Key Technologies Developement and Application of Piloted Autonomous Driving Trucks Project, and the Shanghai Rising Star Program (No. 21QC1400900), and the National Key Research and Development Program of China (No.2016YFB0100901).

{\small
\bibliographystyle{ieee_fullname}
\bibliography{egbib}
}

\end{document}